\documentclass[10pt,twocolumn,letterpaper]{article}

\usepackage{cvpr}
\usepackage{times}
\usepackage{epsfig}
\usepackage{graphicx}
\usepackage{amsmath}
\usepackage{amssymb}

\usepackage{algorithm}
\usepackage{algcompatible}
\algblockdefx{FORP}{ENDFORP}[1]%
  {\textbf{for}#1 \textbf{do in parallel}}%
  {\textbf{end for}}
\usepackage[subrefformat=parens]{subcaption}

\usepackage[pagebackref=true,breaklinks=true,letterpaper=true,colorlinks,bookmarks=false]{hyperref}

\makeatletter
\newcommand{\printfnsymbol}[1]{%
  \textsuperscript{\@fnsymbol{#1}}%
}
\newcommand{\aggregate}[2]{\underset{#2}{\operatornamewithlimits{#1\ }}}
\makeatother

\cvprfinalcopy 
\def\cvprPaperID{****} 

\ifcvprfinal\fi
\begin{document}


\title{Large-scale Landmark Retrieval/Recognition under a Noisy and Diverse Dataset}
\author{Kohei Ozaki\thanks{equal contribution}\\
Recruit Technologies\\
{\tt\small eowner@gmail.com}
\and
Shuhei Yokoo\printfnsymbol{1}\\
University of Tsukuba\\
{\tt\small yokoo@cvlab.cs.tsukuba.ac.jp}
}

\maketitle

\begin{abstract}
The Google-Landmarks-v2 dataset is the biggest worldwide landmarks dataset characterized by a large magnitude of noisiness and diversity.
We present a novel landmark retrieval/recognition system, robust to a noisy and diverse dataset, by our team, smlyaka.
Our approach is based on deep convolutional neural networks with metric learning, trained by cosine-softmax based losses.
Deep metric learning methods are usually sensitive to noise, and it could hinder to learn a reliable metric. To address this issue, we develop an automated data cleaning system.
Besides, we devise a discriminative re-ranking method to address the diversity of the dataset for landmark retrieval.
Using our methods, we achieved 1st place in the Google Landmark Retrieval 2019 challenge and
3rd place in the Google Landmark Recognition 2019 challenge on Kaggle.
\end{abstract}

\section{Introduction}
This paper presents details of our 1st place solution to the Google Landmark Retrieval 2019 challenge~\footnote{https://www.kaggle.com/c/landmark-retrieval-2019/} and 3rd place solution to the Google Landmark Recognition 2019 challenge~\footnote{https://www.kaggle.com/c/landmark-recognition-2019/}.
These two competitions are run parallelly on a large-scale landmarks dataset provided by Google.
The goal of the retrieval challenge is to retrieve all database images which depict the same landmark as a query image, while the goal of the recognition challenge is to recognize a landmark presented in a query image.

The task of retrieval challenge can be viewed as an instance-level image retrieval problem, and the task of recognition challenge can be viewed as an instance-level visual recognition problem.
Both are essential and fundamental problems for academic research and industrial applications.
Recently, deep convolutional neural network based approaches~\cite{EndToEndDIR2017,Radenovic2018FinetuningCI,iccvNohASWH17} have shown astonishing results in these problems.
In our solution, we leverage modern convolutional neural network architectures following this trend.
Besides, cosine-softmax based losses~\cite{ArcFace2018,CosFace2018}, which were initially introduced in face recognition problems, were employed to learn image representations.

The main difficulty of the competitions is that the dataset is quite noisy.
To tackle this challenge, we introduce an automated data cleaning by utilizing a local feature matching method.
In addition, we devise a discriminative re-ranking method leveraging the train set to address the diversity of the dataset for landmark retrieval.

\section{Dataset}
\label{sec:dataset}
The Google-Landmarks-v2~\footnote{https://github.com/cvdfoundation/google-landmark} dataset used for the Google Landmark Retrieval 2019 challenge and the Recognition 2019 challenge
is the largest worldwide landmark recognition dataset available at the time.
This dataset includes over 5M images of more than 200k different landmarks.
It is divided into three sets: train, test, and index. Only samples from the train set are labeled.
The retrieval track asks us to find an image of the same instance (landmark) from the index set,
while the recognition track asks us to answer the corresponding label defined in the train set.
We also used the first version of Google-Landmarks dataset, Google-Landmarks-v1~\cite{iccvNohASWH17}.
The v1 dataset was released after an automated data cleaning step, while the v2 dataset is the raw data.
Since the v2 dataset was constructed by mining web landmark images without any cleaning step, each category may contain quite diverse samples:
for example, images from a museum may contain outdoor images showing the building and indoor images depicting a statue located in the museum.

\subsection{Automated Data Cleaning}
To build a clean train set, we apply spatial verification~\cite{Philbin07} to filtered images by $k$ nearest neighbor search.
Specifically, cleaning the train set consists of a three-step process.
First, for each image representation $x_i$ in the train set, we get its $k$ nearest neighbors from the train set.
This image representation is learned from the v1 dataset.
Second, spatial verification is performed on up to the 100 nearest neighbors assigned to the same label with $x_i$.
This step is necessary to reduce the computational cost.
Finally, if the count of verified images in the second step is greater than the threshold ($t_{\text{frequency}}$),
$x_i$ is added to the cleaned dataset.
In spatial verification, we use RANSAC~\cite{Fischler:1981:RSC:358669.358692} with affine transformation and the deep local attentive features (DELF)~\cite{iccvNohASWH17}, where the threshold of inlier-count is set to 30.
We set $t_{\text{frequency}} = 2$, $k = 1000$ in our experiment.

Table~\ref{tab:dataset_chara} summarizes the statistics of the dataset used in our experiments.
We show the effectiveness of using our cleaned dataset through our experiments in the following sections.

\begin{table}[t]
\begin{center}
\begin{tabular}{|l|r|r|}
\hline
Dataset & \# Samples & \# Labels \\
\hline\hline
Google-Landmarks-v1 & 1,225,029 & 14,951 \\
Google-Landmarks-v2 & 4,132,914 & 203,094 \\
Our clean train set & 1,920,676 & 104,912 \\
\hline
\end{tabular}
\end{center}
\caption{Dataset statistics used in our experiments. The index and test images are not included.}
\label{tab:dataset_chara}
\end{table}

\section{Modeling}
\label{sec:modeling}
To obtain landmark image representation, convolutional neural networks are employed through our pipeline both in the recognition track and the retrieval track.
We use FishNet-150~\cite{FishNet2018}, ResNet-101~\cite{resnet} and SE-ResNeXt-101~\cite{SEResNeXt} as backbones trained with cosine-softmax based losses.
These backbone models are pretrained on ImageNet~\cite{imagenet} and v1 dataset~\cite{iccvNohASWH17} first, and then trained on our cleaned dataset.
On top of that, cosine-softmax based losses were used. Cosine-softmax based losses have achieved impressive results in face recognition.
In this work, we use ArcFace~\cite{ArcFace2018} and CosFace~\cite{CosFace2018} with a margin of 0.3.

We use generalized mean-pooling (GeM)~\cite{Radenovic2018FinetuningCI} for pooling method since it has superior performance than other pooling methods, such as regional max-pooling (R-MAC)~\cite{RMAC} and Compact Bilinear Pooling~\cite{CBP} on our experimental results. $p$ of GeM is set to 3.0 and fixed during the training.


Reduction of a descriptor dimension is crucial since it dramatically affects the computational budget and alleviates risk of over-fitting.
We reduce the dimension to 512 by adding a fully-connected layer after a pooling layer.
Additionally, one-dimensional Batch Normalization~\cite{bn} follows the fully-connected layer to enhance generalization ability.

{\bf Training settings.}
Our implementation is based on PyTorch~\cite{Pytorch}, and four NVIDIA Tesla-V100 GPUs are used for training.
Model training is done by using the stochastic gradient descent
with momentum, where initial learning rate, momentum, weight decay, and batch size are set to 0.001, 0.9, 0.00001, and 32, respectively.
For learning rate scheduler, cosine annealing~\cite{SGDR2017CosineAnnealing} is used.

We employ two different training strategy, one is 5 epochs of training with ``soft'' data augmentation, and the other is 7 epochs of training with ``hard'' data augmentation.
``Soft'' data augmentation includes random cropping and scaling.
``Hard'' data augmentation includes random brightness shift, random sheer translation, random cropping, and scaling.

When constructing the mini-batches for training, gathered images are resized to the same size to feed into networks simultaneously for efficiency.
This mini-batch construction might cause distortions to the input images, degrading the accuracy of the network. 
To avoid this, we choose mini-batch samples so that they have similar aspect ratios, and resize them to a particular size.
The size is determined by selecting width and height from [352, 384, 448, 512] depending on their aspect ratio.
On the final epoch of training, the scale of the input images is enlarged and all batch normalization layers are freezed.
Specifically, the width and height for resizing are chosen from [544, 608, 704, 800] instead of [352, 384, 448, 512].
This approach is beneficial for the network because it enables to exploit more detailed spatial information during training.

{\bf Ensemble.}
In our pipeline, an ensemble is done by concatenating descriptors from different models.
We have six models in total, and each outputs a descriptor with 512 dimensions; Therefore the concatenated descriptor becomes a dimension of 3072.
At the inference time, multi-scale representation is used~\cite{Radenovic2018FinetuningCI}.
We resize the input image at three scale factors of [0.75, 1.0, 1.25], then feed them to the network, and finally average their descriptors.


\section{Retrieval Track}
\label{sec:retrieval}

In this section, we present our pipeline for the Landmark Retrieval Challenge.
Following the conventional approaches based on deep convolutional neural network~\cite{EndToEndDIR2017,Radenovic2018FinetuningCI}, similarity search is conducted by a brute-force euclidean search with L2-normalized descriptors learned by networks.
Besides, to improve retrieval results further, we propose a discriminative re-ranking method which leverages the train set.

\begin{table*}[t]
\begin{center}
\begin{tabular}{|l|l|l|cc|cc|cc|}
\hline
 & & & \multicolumn{2}{|c|}{LB (mAP@100)} & \multicolumn{2}{|c|}{$\mathcal{R}$Oxford (mAP)} & \multicolumn{2}{|c|}{$\mathcal{R}$Paris (mAP)} \\
Backbone & Loss & DA    & Public & Private & Medium & Hard & Medium & Hard \\ \hline\hline
FishNet-150 \cite{FishNet2018} & ArcFace \cite{ArcFace2018} & soft & 28.66   & 30.76    & 80.20   & {\bf 65.70}  & 89.56  & 78.58 \\
FishNet-150 & ArcFace & hard & 29.17  & 31.26     & 80.97  & 64.37 & 89.43  & 78.84 \\
FishNet-150 & CosFace \cite{CosFace2018} & soft & 29.04  & 31.56    & {\bf 82.82}  & 64.82 & 89.64  & 79.07 \\
ResNet-101 \cite{resnet} & ArcFace & hard & 28.57  & 31.07    & 81.18  & 62.62 & 88.56  & 77.63 \\
SE-ResNeXt-101 \cite{SEResNeXt} & ArcFace & hard & {\bf 29.60}  & 31.52    & 80.11  & 61.82 & {\bf 90.22}  & {\bf 79.57} \\
SE-ResNeXt-101 & CosFace & hard & 29.42  & {\bf 31.80}    & 81.11  & 63.14 & 89.07  & 76.97 \\ \hline
\multicolumn{3}{|l|}{Ensemble} & {\bf 30.95}  & {\bf 33.01}    & {\bf 83.42}  & {\bf 66.95} & {\bf 91.80}   & {\bf 82.98} \\ \hline
\end{tabular}
\end{center}
\caption{Performance (mAP [\%]) of our models on the leader-board (LB) of Retrieval track (Public/Private), $\mathcal{R}$Oxford and $\mathcal{R}$Oxford using Medium and Hard evaluation protocols~\cite{RITAC18}. ``DA'' represents the data augmentation strategy (described in Section~\ref{sec:modeling}).}
\label{tab:retrieval_results}
\end{table*}

\subsection{Discriminative Re-ranking}
As described in Section~\ref{sec:dataset}, each landmark category in the dataset may contain diverse samples, such as outdoor and indoor images.
These images are extremely hard to identify as the same landmark without any context.
Thus, these cannot be retrieved as positive samples by a euclidean search only, because they are visually dissimilar and distant in descriptor space.
To solve this, we devise a discriminative re-ranking method exploiting the label information from the train set.

\begin{figure}[t]
\centering
  \begin{minipage}[b]{\hsize}
    \centering
    \includegraphics[width=0.9\linewidth]{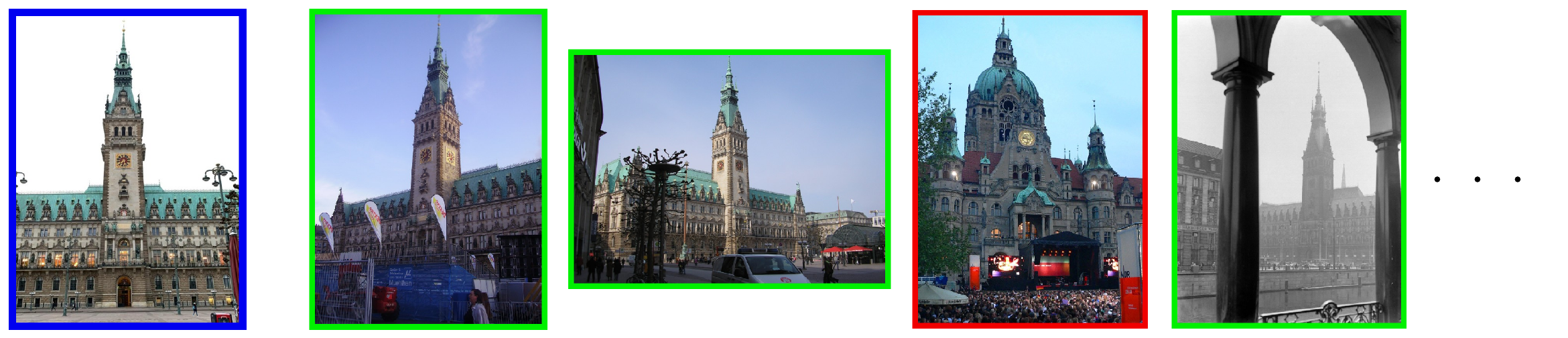}
    \subcaption{A query from the test set (left-most) and original retrieved samples from the index set. Samples to the left are considered more relevant to the query.}
    \vspace{3mm}
    \label{fig:reranking1}
  \end{minipage}
  \begin{minipage}[b]{\hsize}
    \centering
    \includegraphics[width=0.9\linewidth]{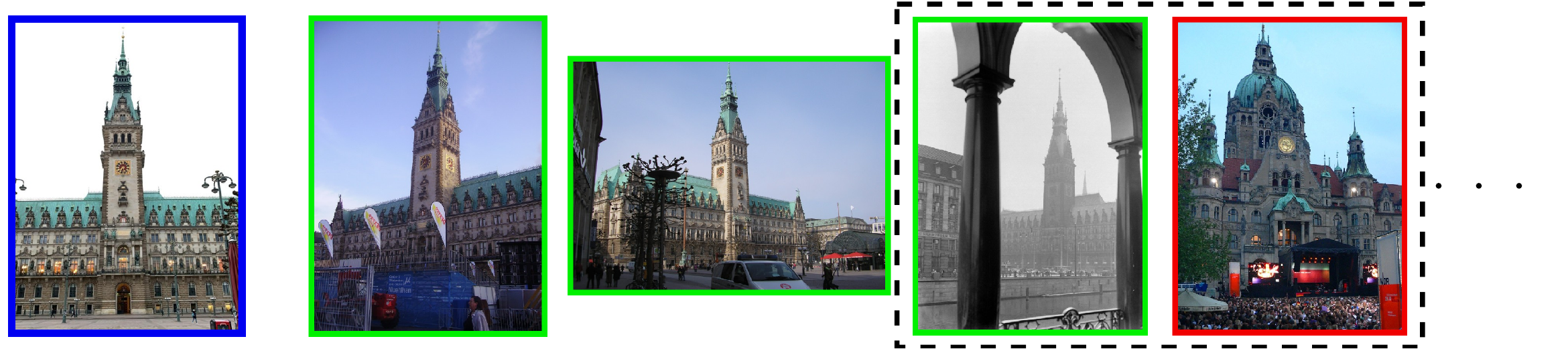}
    \subcaption{More positive samples are moved to the left of negative samples.}
    \vspace{3mm}
    \label{fig:reranking2}
  \end{minipage}
  \begin{minipage}[b]{\hsize}
    \centering
    \includegraphics[width=1.0\linewidth]{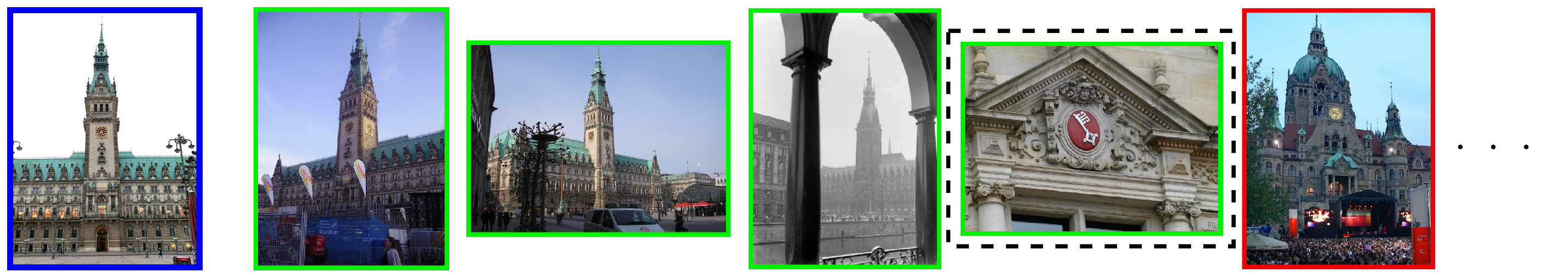}
    \subcaption{
    Positive samples from the entire index set that were not retrieved by the similarity search are appended after the re-ranked positive samples.
    }
    \label{fig:reranking3}
  \end{minipage}
  \caption{
  Our re-ranking procedure.
  ``Positive samples'' represents the samples which predicted same landmark as a predicted landmark of a test sample.
  ``Negative samples'' represents the samples which predicted different landmark from a predicted landmark of a test sample.
  Query images are in blue, positive samples are in green and negative samples are in red.
  }
  \label{fig:reranking}
\end{figure}

As a first step, we predict a landmark-id of each sample from the test set and index set following the same procedure as in the recognition track (Section~\ref{sec:recognition}) before re-ranking.
We treat the index set samples that are predicted the same landmark of each test set sample as ``positive samples''.
Likewise, we treat the index set samples that are predicted the different landmark of each test sample as ``negative samples''.
Figure~\ref{fig:reranking} illustrates a procedure of our re-ranking method performed on actual examples from the dataset.
Figure~\ref{fig:reranking1} shows a query from the test set (in blue) and retrieved samples from the index set by similarity search (in green for positive samples and red for negative samples).
Here, we consider images to the left to be more relevant than the ones to the right. 
Therefore, the right-most positive sample is considered less irrelevant than the negative sample on its left due to several factors (e.g., occlusion).
It is desirable to ignore such trivial conditions for retrieval landmarks. 
In Figure~\ref{fig:reranking2}, positive samples are moved to the left of the negative samples in the ranking.
This re-ranking step can make results more reliable, becoming less dependent on those trivial conditions.
Finally, we append positive samples from the entire index set, which are not retrieved by the similarity search, after the re-ranked positive samples~(Figure~\ref{fig:reranking3}).
This step enables to retrieve visually dissimilar samples to a query by utilizing the label information of the train set.

Our re-ranking system is related to Discriminative Query Expansion~\cite{Arandjelovic2012DiscriminativeQueryExpansion}, but our system doesn't require to train discriminative models.

\subsection{Results}

We evaluate our models on two landmark retrieval benchmarks, $\mathcal{R}$Oxford5k and $\mathcal{R}$Paris6k~\cite{RITAC18}.
Also, we evaluate them on the Google Landmark Retrieval 2019 challenge.
Results are presented in Table~\ref{tab:retrieval_results}.
We use the FAISS library~\cite{JDH17}, an efficient implementation of euclidean search, for all experiments.

On the Google Landmark Retrieval 2019 challenge,
training on the clean train set significantly boosts our score compared to only training on the train set from the v1 dataset.
Without training on the clean train set (only training on the train set from the v1 dataset), our best single model scores 19.05/20.99 on the public/private set respectively.
We also tried database-aside feature augmentation (DBA)~\cite{Arandjelovic2012DiscriminativeQueryExpansion}, alpha-Query expansion (QE)~\cite{Radenovic2018FinetuningCI}, Yang's diffusion~\cite{Yang2019EfficientIR} and Graph traversal~\cite{EGT2019}, but their performance improvement were limited.

\begin{table}[t]
\begin{center}
\begin{tabular}{|l|c|c|}
\hline
Method & Public & Private \\
\hline\hline
Single best model & 29.42 & 31.80 \\
+ Ensemble 6 models & 30.95 & 33.01 \\
+ DBA, QE & 31.41 & 32.81 \\
+ Our Re-ranking & 35.69 & 37.23 \\
\hline
\end{tabular}
\end{center}
\caption{Performance of our pipeline on the public set and the private set from the retrieval track leader-board. mAP@100~[\%] is used for evaluation.}
\label{tab:retrieval_pipeline_results}
\end{table}

Table~\ref{tab:retrieval_pipeline_results} shows the performance improvement by our ensemble model and re-ranking method.
Without any re-ranking method, such as our discriminative re-ranking, the performance of our ensemble model is equivalent to 3rd place on the Landmark Retrieval 2019 challenge.

\section{Recognition Track}
\label{sec:recognition}
In this section, we present our pipeline for the Landmark Recognition Challenge.
Our pipeline consists of three steps: euclidean search of image representation, soft-voting, and post-processing.

The first step is to find the top-$k$ neighborhoods in the train set by a brute-force euclidean search.
Secondly, the label of the given query is estimated by the majority vote of soft-voting based on the sum of cosine similarity between the neighboring training image and the query. The sum of cosine similarity is used as a confidence score.
Finally, we suppress the influence of distractors by a heuristic approach.

\subsection{Soft-voting with spatial verification}
To make the score of similarity more robust,
we used RANSAC and inlier-based methods for scoring confidence.
They are widely used in retrieval methods as a post-processing step to reduce false positives.

Let us have a set of $q$'s neighbors in image representation $\textbf{N}_l(q)$, where its members are assigned to the label $l$.
Let $R(x_i, x_j)$ be the inlier-count between two image representation $x_i$ and $x_j$ in RANSAC.
For a given image representation of query $q$, we estimate its label $y(q)$ as follows:
\begin{align*}
  y(q) &= \aggregate{argmax}{l} s_l, \\
  s_l &= \sum_{i\in \textbf{N}_l (q)} (1 - ||x_i - q||^2) + \min(t, R(x_i, q)) / t,
\end{align*}
where $t$ is the threshold parameter to verify the matching. We set $t=70$ in our experiment.
We used 3 nearest neighbors in descriptor space for soft-voting.

\subsection{Post-processing for distractor}
The metric of the recognition track, Global Average Precision (GAP)~\cite{PerronninCVPR09}, penalizes if non-landmark images (distractors) are predicted with higher confidence score than landmark images.
Hence, it is essential to suppress the prediction confidence score of these distractors.

We observed that categories frequently predicted in the test set are likely non-landmark (e.g., flowers, portraits, and airplanes).
From this observation, we treat categories that appear more frequently than 30 times in the test set as non-landmark categories.
Each confidence score of these categories is replaced with multiplying their frequency by -1 to suppress them.

\subsection{Results}

\begin{table}[t]
\begin{center}
\begin{tabular}{|l|c|c|}
\hline
Method & Public & Private \\
\hline\hline
Signle best model & 0.1872 & 0.2079 \\
+ Spatial verification~\cite{Philbin07} & 0.2911 & 0.3373 \\  
+ Ensemble 6 models & 0.2966 & 0.3513 \\  
+ Post-processing & 0.3066 & 0.3630 \\ 
\hline
\end{tabular}
\end{center}
\caption{Performance of our pipeline on the public set and the private set from the recognition track leader-board. GAP is used for evaluation.}
\label{table:recog19results}
\end{table}

We show the results of our pipeline in Table~\ref{table:recog19results}.
Both cases with and without spatial verification were evaluated.
The GAP score is significantly improved by the ensemble and spatial verification.
Our post-processing step also helps to improve the evaluation score.
In Landmark Recognition 2019 challenge, our pipeline won the 3rd place~\footnote{The best score described in Table~\ref{table:recog19results} is equivalent to 2nd place in Landmark Recognition 2019 challenge. Our final submission added DBA step and it degraded our best score.}.

\section{Conclusion}

In this paper, we presented a large-scale landmark retrieval and recognition system by team smlyaka.
Our experimental results show that our automated data cleaning and discriminative re-ranking play an important role in the noisy and diverse dataset.

\section*{Acknowledgements}
Computational resource of AI Bridging Cloud Infrastructure (ABCI) provided by National Institute of Advanced Industrial Science and Technology (AIST) was used.
We would like to thank Erica Kido Shimomoto for reviewing of our paper.

{\small
\bibliographystyle{ieee}
\bibliography{egbib}
}

\end{document}